# State of the Art Optical Character Recognition of 19th Century Fraktur Scripts using Open Source Engines

Christian Reul, Uwe Springmann, Christoph Wick, and Frank Puppe

Chair for Artificial Intelligence and Applied Computer Science, University of Würzburg

**Abstract:**
In this paper we evaluate Optical Character Recognition (OCR) of 19th century Fraktur scripts without book-specific training using mixed models, i.e. models trained to recognize a variety of fonts and typesets from previously unseen sources. We describe the training process leading to strong mixed OCR models and compare them to freely available models of the popular open source engines OCRopus and Tesseract as well as the commercial state of the art system ABBYY. For evaluation, we use a varied collection of unseen data from books, journals, and a dictionary from the 19th century. The experiments show that training mixed models with real data is superior to training with synthetic data and that the novel OCR engine Calamari outperforms the other engines considerably, on average reducing ABBYYs character error rate (CER) by over 70%, resulting in an average CER below 1%.

## 1   Introduction

During the last few years, great progress has been made on OCR methods which can mainly be attributed to the introduction of a line based recognition approach using recurrent neural networks (Breuel et al. 2013). Since this breakthrough, impressive recognition accuracies beyond 98% have been achieved on a variety of materials, ranging from the earliest printed books (Springmann et al. 2016; Springmann and Lüdeling 2017) to modern prints (Breuel 2017; Wick et al. 2018). Early prints show a high variability in terms of printing types and therefore usually require book-specific training in order to reach desirable character error rates (CER) below 1-2%. On the contrary, modern typography is much more regular and mixed models, i.e. models trained on a variety of fonts and typesets from different sources, comfortably achieve CERs well below 1% without any book-specific training. Apart from the aforementioned introduction of new recognition techniques and network structures, several methodical improvements like pretraining (transfer learning) and majority or confidence voting have been introduced and successfully evaluated, especially for the application on early printed books (Reul et al. 2018).

Printings from the 19[th] century represent a middle ground between the two periods introduced above, considering both the variability of typesets and the state of preservation of the scans. Mixed models have achieved encouraging results without the need for book-specific training but the expectable recognition accuracy still is substantially lower than for prints from the 21[st] century (Breuel et al. 2013). Just as for modern prints, there is a great need for highly performant mixed models for 19[th] Fraktur scripts since there are masses of scanned data available online, consisting of a variety of materials including novels, newspapers, journals, and even dictionaries.

In this paper, we describe the training procedure leading to our own strong mixed models and compare the evaluation results to those achieved by other main OCR engines and their respective models on a variety of Fraktur scripts. In particular, we report results from OCRopus, Tesseract, and ABBYY Finereader each with their own standard Fraktur model as well as OCRopus and Calamari with a mixed model trained on a Fraktur corpus of the 19[th] century.

## 2   Related Work

Only few evaluation results are available on 19th century Fraktur OCR data. A rare exception is the evaluation of the Fraktur model of OCRopus trained on around 20,000 mostly synthetically generated text lines (Breuel et al. 2013). Evaluation on two books of different scan qualities yielded impressive CERs of 0.15% and 1.37% respectively. There exist other evaluations on more recent (Breuel et al. 2013) or older texts (Springmann and Lüdeling 2017) yielding better and worse results, respectively. An evaluation of OCR data on a wider range of Fraktur texts of different quality is missing.

## 3   Methods

In this section we briefly describe the OCR engines ABBYY Finereader, OCRopus, Tesseract, and Calamari, our training and evaluation data as well as the transcription guidelines.

### 3.1   OCR Engines

For contemporary material the proprietary ABBYY OCR engine (https://www.ABBYY.com) clearly defines the state of the art for layout analysis and OCR covering close to 200 recognition languages including Fraktur printed in the 18-20th centuries with an "Old German" dictionary which we used for our experiments.

The open source engine OCRopus was the first one to implement the pioneering line based approach introduced by Breuel et al. (Breuel et al. 2013) using bidirectional LSTM networks. Apart from the superior recognition capabilities compared to glyph-based approaches, this method has the advantage of allowing the user to train new models very comfortably by just providing image/text pairs on line level.

Calamari (https://github.com/Calamari-OCR), also available under an open source license, implements a deep CNN-LSTM network structure instead of the shallow LSTM used by OCRopus. It yields superior recognition capabilities compared to OCRopus and Tesseract (Wick et al. 2018). Because of its Tensorflow backend it is possible to utilize GPUs in order to support very fast training and recognition. In addition, it supports the training of voting ensembles and pretraining, i.e. it uses an already existing model as a starting point instead of training from scratch.

Until recently, the open source OCR engine Tesseract (https://github.com/tesseract-ocr) used individual glyphs rather than entire text lines for training and recognition. However, version 4.0 alpha also added a new OCR engine based on LSTM neural networks and a wide variety of trained mixed models. Like ABBYY and contrary to OCRopus and Calamari, Tesseract supports the use of dictionaries and language modelling.

### 3.2   Training Data

To achieve high quality results on early prints it is usually necessary to perform a book-specific training. For our 19th century mixed model we try to avoid this by training on a wide variety of sources over four subsequent training steps (see Table 1). First, we use corpora with texts from different centuries for pretraining to achieve a certain overall robustness. Next, the training continues by incorporating synthetic data generated from freely available Fraktur fonts. The training concludes with the addition of real Fraktur data from the 19th century. After training on the entire data set, we perform a final refinement step in which we only use a subset of at most 50 lines per book in order to prevent the model from overfitting to the books with a high number GT lines available (10,000+ compared to less than 50 for some books). The described data are mostly available online in the GT4HistOCR corpus (Springmann et al. 2018).

*Table 1. Corpora used for training our mixed models. Apart from the data available in the GT4HistOCR corpus we also incorporated lines from the Archiscribe project (https://archiscribe.jbaiter.de) and the GitHub repository of Jesper Zedlitz (JZE, https://github.com/jze/ocropus-model_fraktur).*

| Data | Cent. | # Books | # Lines | Lang | Step |
|---|---|---|---|---|---|
| ENHG | 15 | 9 | 24,766 | ger | Pretraining |
| Kallimachos | 15,16 | 9 | 20,929 | ger, lat | Pretraining |
| EML | 15-17 | 12 | 10,288 | lat | Pretraining |
| RIDGES | 15-19 | 20 | 13,248 | ger | Pretraining |
| UW3 | 20 | - | 96,481 | eng | Pretraining |
| Synth. | - | 66 fonts | 99,214 | ger | Synth. Data |
| DTA19 | 19 | 39 | 243,942 | ger | Real Data |
| Archiscribe | 19 | 103 | 3,430 | ger | Real Data |
| JZE | 19 | 8 | 1,636 | ger | Real Data |
| DTA19 | 19 | 39 | 1,950 | ger | Refinement |
| Archiscribe | 19 | 103 | 3,429 | ger | Refinement |
| JZE | 19 | 8 | 355 | ger | Refinement |

### 3.3 Evaluation Data

For evaluation, we used four corpora from the 19th century (Table 2, top), which were completely different from the training data, and consisted of 20 different evaluation sets (Table 2, bottom).

*Table 2. "Novels" (N) is a corpus consisting of novels currently collected and captured by the Chair for Literary Computing and German Literary History of the University of Würzburg. The "OCR-Testset" (O, https://github.com/cisocrgroup/Resources/tree/master/ocrtestset) consists of a novel and a journal. "Daheim" comprises four volumes of a German journal and "Sanders" (S) is a German dictionary provided by the Berlin-Brandenburg Academy of Sciences and Humanities.*

| Data | Period | # Lines | # Books |
|---|---|---|---|
| Novels | 1781-1873 | 3,483 | 13 |
| OCR-TS | 1809-1841 | 465 | 2 |
| Daheim | 1865-1875 | 583 | 4 vol. |
| Sanders | 1865 | 630 | 1 |

| ID | (Short) Title | # Lines |
|---|---|---|
| N-1781 | Eleonore | 305 |
| N-1803 | Liebe-Hütten | 184 |
| N-1810 | Der Held des Nordens | 264 |
| N-1818 | Reinhold | 253 |
| N-1826 | Frauenwürde | 268 |
| N-1836 | Die Ruinen im Schwarzwalde | 318 |
| N-1848 | Levin | 269 |
| N-1851 | Georg Volker | 264 |
| N-1859 | Der beseelte Schatten | 260 |
| N-1865 | Gefahrvolle Wege | 333 |
| N-1869 | Der Arzt der Seele | 250 |
| N-1870 | Die Bank des Verderbens | 273 |
| N-1873 | Natürliche Magie | 242 |
| O-1809 | Wahlverwandtschaften | 223 |
| O-1841 | Grenzboten | 242 |
| D-1865 | Daheim volume 1865 | 134 |
| D-1875 | Daheim volume 1875 | 144 |
| D-1882 | Daheim volume 1882 | 142 |
| D-1892 | Daheim volume 1892 | 163 |
| S-1865 | Sanders Dictionary | 630 |

Figure 1 shows some example lines.

*Figure 1. Example line images of the 20 evaluation works in the order given in Table 3. For practical reasons, all lines have been vertically normalized and some lines have been shortened.*

### 3.4 Transcription Guidelines and Resulting Codec

Before starting the training, we had to make several decisions regarding the codec, i.e. the set of characters known to the final model. We kept the long s, resolved all ligatures with the exception of ß (sz), regularized Umlauts like å, ô, ů, quotation marks, different length hyphens, the r rotunda (ꝛ ) and mapped the capital letters I and J to J. Applying these rules resulted in a codec consisting of 93 characters:

- special characters: ␣!"\&'()[]*,-./:;=?§⸗
- digits: 0123456789
- lower case letters: abcdefghijklmnopqrsſßtuvwxyz
- upper case letters: ABCDEFGHJKLMNOPQRSTUVWXYZ
- characters with diacritica: ÄÖÜäöüàèé

## 4 Evaluation

Table 3 summarizes the results of applying the four OCR-Engines to the 20 data sets from Table 2. For all evaluations the experiments were performed on well segmented line images provided by ABBYY.

*Table 3. CERs in percent of different OCR engines and their respective mixed models: Tesseracts "frk_best" model (Tess), OCRopus with its standard Fraktur model (FRK) and the mixed model trained by us (OCRo), and Calamari with and without voting.*

| Data | Tess single | FRK single | OCRo single | Abbyy default | Calamari single | Calamari voted |
|---|---|---|---|---|---|---|
| N-1781 | 6.61 | 4.08 | 2.48 | 2.79 | 0.81 | **0.56** |
| N-1803 | 17.17 | 18.21 | 11.30 | 26.54 | 6.38 | **4.75** |
| N-1810 | 5.26 | 5.30 | 1.92 | 3.22 | 0.45 | **0.21** |
| N-1818 | 7.90 | 7.73 | 3.85 | 9.30 | 1.85 | **0.96** |
| N-1826 | 2.77 | 1.00 | 0.40 | 1.04 | 0.08 | **0.01** |
| N-1836 | 6.88 | 4.68 | 2.01 | 2.70 | 0.70 | **0.56** |
| N-1848 | 1.58 | 1.17 | 0.33 | 0.57 | 0.08 | **0.02** |
| N-1851 | 1.93 | 0.63 | 0.24 | 0.70 | 0.09 | **0.04** |
| N-1855 | 4.58 | 4.42 | 1.38 | 3.83 | 0.80 | **0.58** |
| N-1859 | 2.19 | 1.42 | 0.31 | 0.38 | 0.17 | **0.08** |
| N-1865 | 2.44 | 1.31 | 0.62 | 1.23 | 0.19 | **0.13** |
| N-1870 | 2.09 | 1.97 | 0.43 | 0.47 | 0.26 | **0.10** |
| N-1873 | 2.53 | 1.14 | 0.32 | 0.34 | 0.22 | **0.14** |
| N-all | 4.39 | 3.42 | 1.58 | 3.13 | 0.71 | **0.47** |
| O-1809 | 3.04 | 2.22 | 1.13 | 1.62 | 0.26 | **0.20** |
| O-1841 | 2.09 | 1.06 | 0.60 | 0.79 | 0.13 | **0.07** |
| O-all | 2.40 | 1.44 | 0.77 | 1.06 | 0.17 | **0.11** |
| D-1865 | 2.10 | 1.85 | 0.71 | **0.16** | 0.26 | 0.17 |
| D-1875 | 1.50 | 0.85 | 0.17 | **0.04** | 0.09 | 0.09 |
| D-1882 | 1.53 | 1.17 | 0.43 | **0.09** | 0.20 | 0.12 |
| D-1892 | 0.90 | 0.45 | 0.23 | **0.01** | 0.02 | 0.01 |
| D-all | 1.48 | 1.05 | 0.38 | **0.07** | 0.17 | 0.09 |
| S-1865 | 5.12 | 10.02 | 5.91 | 5.47 | 2.74 | **2.14** |
| NOD | 3.68 | 2.80 | 1.29 | 2.38 | 0.55 | **0.37** |
| All | 3.87 | 3.76 | 1.90 | 2.80 | 0.84 | **0.61** |

## 5  Discussion

A striking result is the great variation among the CERs, e.g. by a factor of more than 2,500 from 26.54% to 0.01% for ABBYY and more than 400 from 4.75% to 0.01% for Calamari voted, which probably depends on the quality of the scans as well as the similarity of each font to the training data. Furthermore, training a model on real Fraktur data outperforms a model trained on mostly synthetic data generated for Fraktur (e.g. FRK vs. OCRo). The self-trained Calamari models achieve the best results, outperforming ABBYY by 70% without voting and even by 78% with voting averaged over all 20 datasets yielding an average CER below 1%.

For all approaches, the most frequent error either consists in the insertion (Tesseract) or the deletion of whitespaces (all others) leading to merged or splitted words. This represents a common problem with historical prints, as the inter word distances vary heavily. The error distribution varies considerably for the different engines. For example, in the case of ABBYY the three most frequent errors make up to less than 5% of all errors, whereas OCRopus (close to 9%) and Calamari (over 15%) show a considerably more top-heavy distribution.

# 6   Conclusion and Future Work

Our evaluations showed that open source engines can outperform the commercial state-of-the-art system ABBYY by up to 78% if properly trained. The resulting models as well as the data required to adjust the models codec are publicly available (https://github.com/chreul/19th-century-fraktur-OCR). Further improvements can be expected by providing more ground truth for training the mixed model and by using even deeper neural networks than the Calamari default. While ABBYY already has strong post processing techniques available, this represents an opportunity to improve the results achieved by Calamari and OCRopus even further, in particular the inclusion of dictionaries and language models.